\documentclass[11pt]{article}
\usepackage{nodalida2025}
\usepackage{times}
\usepackage{tabularx}
\usepackage{graphics}
\usepackage{booktabs}
\usepackage{array}
\usepackage{url}
\usepackage{hyperref}
\usepackage{latexsym}
\usepackage{svg}
\usepackage{comment}
\usepackage[utf8]{inputenc}
\usepackage[T2A,T1]{fontenc}
\usepackage{csquotes}
\MakeOuterQuote{"}

\usepackage{textcomp}

\aclfinalcopy 

\title{FinerWeb-10BT: Refining Web Data with LLM-Based Line-Level Filtering}

\author{Erik Henriksson* \\
  University of Turku \\
  {\tt erik.henriksson@utu.fi} \\\And
  Otto Tarkka* \\
  University of Turku \\
  {\tt ohitar@utu.fi} \\\And
  Filip Ginter \\
  University of Turku \\
  {\tt figint@utu.fi} \\
  }

\date{\today}

\begin{document}
\maketitle

\begingroup\def\thefootnote{*}\footnotetext{These authors contributed equally.}\endgroup

\begin{abstract}

Data quality is crucial for training Large Language Models (LLMs). Traditional heuristic filters often miss low-quality text or mistakenly remove valuable content. In this paper, we introduce an LLM-based line-level filtering method to enhance training data quality. We use GPT-4o mini to label a 20,000-document sample from FineWeb at the line level, allowing the model to create descriptive labels for low-quality lines. These labels are grouped into nine main categories, and we train a DeBERTa-v3 classifier to scale the filtering to a 10B-token subset of FineWeb. To test the impact of our filtering, we train GPT-2 models on both the original and the filtered datasets. The results show that models trained on the filtered data achieve higher accuracy on the HellaSwag benchmark and reach their performance targets faster, even with up to 25\% less data. This demonstrates that LLM-based line-level filtering can significantly improve data quality and training efficiency for LLMs. We release our quality-annotated dataset, FinerWeb-10BT, and the codebase to support further work in this area.
\end{abstract}

\section{Introduction} \label{sec:introduction}

In recent years, the size of large language models (LLMs) and their training datasets has expanded tremendously, as companies and researchers strive to build increasingly capable models. In fact, if current trends continue, we may run out of human-generated text data within a decade \cite{villalobos_2024}. This has led to a growing interest in data quality over quantity: rather than only expanding datasets, researchers are exploring ways to achieve high performance with smaller, cleaner datasets. Recent studies suggest that removing low-quality text from training data can improve model performance, even when the overall size of the dataset is reduced \citep{longpre2023}.

Furthermore, training state-of-the-art (SOTA) language models requires significant computational resources, which are expensive and, depending on the power source, can contribute to climate change. For example, the carbon emissions from training GPT-3 have been estimated at 552 tCO$_{2}$e \cite{patterson2021}, while Meta reports that training the 405 billion parameter Llama 3.1 emitted 8,930 tCO$_{2}$e \cite{llama3.1_github}. Smaller, but higher quality datasets will speed up training and, thus, high-quality data are necessary to train not only better models but also greener ones.

While several publicly available datasets are used for training LLMs, many recent datasets are still cleaned using simple heuristic filters, which often leave substantial amounts of low-quality text while potentially discarding clean text. Machine-learning techniques offer a promising alternative, as they enable models to identify patterns related to data quality. However, labeling data to train such models is a tedious and time-consuming process. In this paper, we address these issues by investigating the following research questions (RQs):

\begin{description} 
    \item[RQ1:] How well can an LLM identify low-quality content missed by heuristic filters?
    \item[RQ2:] Does LLM-based quality filtering of training datasets improve model performance?
\end{description}

To examine these questions, we analyze FineWeb, a dataset that claims to provide "the finest text data at scale" \cite{fineweb}. Using GPT-4o mini \cite{openai2024gpt4omini}, we label a 20,000-document sample from FineWeb, classifying each line as either \textit{Clean} or belonging to one of several low-quality categories, such as \textit{copyright notice}, \textit{programming code}, or \textit{formatting elements}. Instead of defining a label taxonomy ourselves, we allow the model to generate its own labels as needed, resulting in 547 unique low-quality labels. After refining these labels, we group them into nine broader categories for easier classification. Next, we train a DeBERTa-v3 \cite{he2021debertav3} classifier using the labeled data to scale the filtering process. This classifier allows us to automatically detect low-quality content in a larger 10B-token sample of FineWeb. Finally, we evaluate the impact of LLM-based filtering by training GPT-2 models \cite{Radford2019LanguageMA} on both the filtered and unfiltered datasets. 

We release our quality-annotated dataset, FinerWeb-10BT, available at https://huggingface.co/datasets/TurkuNLP/finerweb-10bt. The code to replicate our experiments is also provided at https://github.com/TurkuNLP/finerweb-10bt.

\section{Background} \label{sec:background}

A recent survey by \citet{surveydataselection} discusses the many steps involved in selecting data for training LLMs, including language filtering, deduplication, removal of toxic or explicit content, and heuristic-based data quality filtering. Our focus here is on the latter two---data filtering and heuristic approaches---using an LLM-driven approach to refine data quality more precisely. As \citet{surveydataselection} note, there is no universal standard for "high-quality" data. In this work, we define it as human-written, continuous English text from the main content of a website, reflecting natural language use across diverse contexts and domains. Examples include core text from interviews, forum posts, news articles, blogs, and recipes. In contrast, low-quality content includes recurring elements like navigational menus, copyright notices, programming code, and metadata.

Given that LLMs require vast amounts of text data for training, the Internet has become a primary source for these data. Since 2008, CommonCrawl has collected a corpus of approximately 10 petabytes of web content \cite{commoncrawl}. Despite its size, CommonCrawl is neither a complete nor fully representative sample of the Internet, but it serves as a foundational source for building refined datasets used in LLM training. Here, we focus on three major datasets sourced from CommonCrawl: C4 \cite{C4}, RefinedWeb \cite{refinedweb}, and FineWeb. These datasets use different preprocessing techniques to filter out unwanted material, each with its strengths and weaknesses. We discuss these datasets because their preprocessing methods are well-documented, which allows us to make meaningful comparisons.

All three datasets extract plaintext from HTML documents. C4 uses the WET files provided by CommonCrawl, which come with pre-extracted plaintext, whereas RefinedWeb and FineWeb use trafilatura\footnote{https://trafilatura.readthedocs.io/en/latest/} to extract text directly from HTML. Although trafilatura and similar tools remove much of the unwanted noise, further preprocessing is often required. For instance, \citet{refinedweb} note that "many documents remain interlaced with undesirable lines" despite using trafilatura. Deduplication and language filtering are also important aspects of document cleaning but we do not focus on them in this paper, as they are specialized techniques not directly related to line-level text quality.

Existing filtering methods can be grouped into three levels based on their precision: document level, line level, and character level. By far the most common method is document level filtering, which removes entire documents based on simple rules. Examples include filtering documents with phrases like "lorem ipsum", documents with fewer than three sentences, or documents with excessive repetition. Line level filtering targets specific lines within documents, removing lines that contain terms like "javascript", consist solely of numbers, or fall below a certain length threshold. Character level filtering is less common and is only applied in one of the three datasets: in C4, citation markers commonly found in Wikipedia, such as "[1]" and "[citation needed]", are removed.

Document level heuristic filtering is efficient for quickly removing large volumes of low-quality data, but it can result in the loss of substantial high-quality text. In contrast, line and character level filtering provide more precision by targeting specific content but they require significantly more computational resources at scale. Simple heuristics, such as removing lines that contain the word "javascript" can be hit or miss, sometimes discarding valuable data along with the low-quality content. Given the vast size of datasets like CommonCrawl, creating a simple filtering system that only removes undesirable content without impacting valuable data is nearly impossible. The filters that are used are also often dataset and language specific. For example, FineWeb applies a heuristic that removes documents where "the fraction of lines shorter than 30 characters is >= 0.67" \cite[p. 7]{fineweb}, but this threshold was determined through extensive manual testing and is specific to that dataset. 

An ideal quality filter would work across languages and datasets, avoiding trial-and-error by focusing on actual text quality rather than proxies like line length or keywords. It should also be efficient, removing only low-quality content while keeping valuable data intact. LLMs bring us closer to this goal: rather than using heuristics, they assess text quality directly, enabling granular filtering, even within mostly clean documents. Since LLMs are effective at producing fluent and readable text, they are likely well suited to identifying high-quality text across different languages and datasets. However, it should be noted that while SOTA LLMs are fluent in English and other high-resource languages, their performance in low-resource languages is consistently worse \citep{li2024languagerankermetricquantifying}. In this study, we only analyze English documents, and care should be taken before generalizing the results to other languages or multilingual datasets.

The use of LLMs for quality filtering is a relatively new approach, and best practices are still emerging. For instance, \citet{llama3herdmodels} utilize Llama 2 to assess the quality of web documents for training Llama 3, but details of their methodology are vague. The recent trend of withholding full training datasets for SOTA models has made it difficult to understand the extent to which LLMs are currently used in data preprocessing \citep{nguyen-etal-2024-culturax, maini-etal-2024-rephrasing}. Other efforts, such as those by \citet{wettig2024}, involve ranking documents based on quality using GPT-3.5, evaluating factors such as style, educational value, and factuality. Similarly, Llama 3 was used to create the FineWedEdu dataset by evaluating educational content quality, and \citet{gunasekar2023textbooksneed} employ GPT-4 to annotate code datasets based on educational value. 

Our approach differs from prior work by focusing on general-purpose data quality improvements rather than curating specialized datasets. We aim to broadly enhance training data quality through LLM-driven filtering that removes low-quality lines with minimal manual intervention. This allows us to assess how automated filtering can improve training data and, ultimately, model performance in foundation model training.

\section{Methods} \label{sec:methods}

Our data source is FineWeb \cite{fineweb}, a 15-trillion-token collection of English text sourced from CommonCrawl and preprocessed with standard heuristics. The preprocessing includes steps such as length thresholds, string matching, language and URL filtering, and deduplication. Despite these measures, the authors of FineWeb acknowledge that the dataset could benefit from further refinement. For more details on the preprocessing steps, see the original paper \cite{fineweb}. In our study, we use a 10B-token (\~15 million documents) sample from FineWeb, FineWeb-10BT\footnote{https://huggingface.co/datasets/HuggingFaceFW/fineweb/\\viewer/sample-10BT}.

Our preprocessing pipeline consists of several steps. First, we use GPT-4o mini \cite{openai2024gpt4omini} to label a sample of 20,000 documents from FineWeb at the line level. The model is tasked with generating descriptive labels for each line, categorizing them as either high-quality (\textit{Clean}) or into low-quality categories. This labeling process is data-driven, allowing the model to create a dynamic labeling scheme rather than relying on predefined categories. Previous research has shown that LLMs can be used to annotate data and create label taxonomies \citep{TnT-LLM}.

Next, we use OpenAI's o1-preview model \cite{openai2024openaio1preview} to group the numerous labels generated by GPT-4o mini into a smaller, more manageable set. This forms the basis of a classification system, which we use to train a small encoder-based classifier. This classifier scales the labeling process by assigning quality scores throughout the FineWeb-10BT dataset, enabling line-level filtering of low-quality content.

To evaluate our filtering, we train GPT-2 models \cite{Radford2019LanguageMA} on both the cleaned and original versions of FineWeb-10BT. We compare model performances using the HellaSwag benchmark \cite{zellers2019hellaswagmachinereallyfinish}, a widely used test for commonsense reasoning in language models. This allows us to assess whether the filtering improves training data quality and model performance.

Given the complexity of Internet text data \citep{laippala_etal}, defining low-quality categories in advance is challenging. Our data-driven approach, by contrast, allows the LLM to dynamically create labels based on the content it encounters, rather than relying on fixed categories. We believe this approach enables a more flexible and detailed analysis of low-quality content in FineWeb compared to rule-based methods or predefined categorizations.

\section{Experiments and results} \label{sec:experiments}

\subsection{Labeling FineWeb using GPT-4o mini} \label{subsec:gpt_labeling}

We begin by labeling a 20,000-document sample from FineWeb-10BT using the GPT-4o mini model. The model is prompted to classify each line as either \textit{Clean} (high quality and suitable for training large language models) or assign a descriptive label if the line contains low-quality content, such as HTML tags or random symbols. Initially, the model generates its own descriptive labels, which are then added to a list for subsequent classification. As the model processes more documents, it selects labels from the existing list or creates new ones if necessary. To avoid bias from label order, the list is shuffled after each iteration.

We split the documents into batches of up to 15 consecutive lines. The model receives a prompt, a list of labels, and a batch of lines. Since the lines are consecutive, each one is evaluated in context, providing the model with more information for accurate labeling. For documents containing a single line longer than 200 characters, the line is split into segments of no more than 200 characters, using sentence-ending punctuation as the split point. This prevents output errors, which we observed when processing excessively long lines during preliminary tests. Segmenting these lines also enables more precise analysis.

This process results in quality labels for 328,472 lines. Of these, 274,343 lines (83\%) are labeled as \textit{Clean}. For low-quality lines, the model generates 547 unique descriptive labels. However, we find that many of these labels are assigned to one line only; in fact, 142 labels appear only once. Upon inspection, we notice many of the lines could be considered high-quality and, thus, to streamline the label set, we map all these infrequent labels to \textit{Clean}. For the remaining labels, we take a sample of lines and manually verify that they represent genuinely low-quality content. If the majority of lines for a particular label are of high quality, we remap that label to \textit{Clean}. After this refinement, the number of descriptive labels is reduced to 382, with 45,205 lines (14\%) classified as low-quality. Conversely, 86\% of the dataset is now labeled \textit{Clean}.

\begin{figure}[!ht]
    \centering
        \includegraphics[width=\columnwidth]{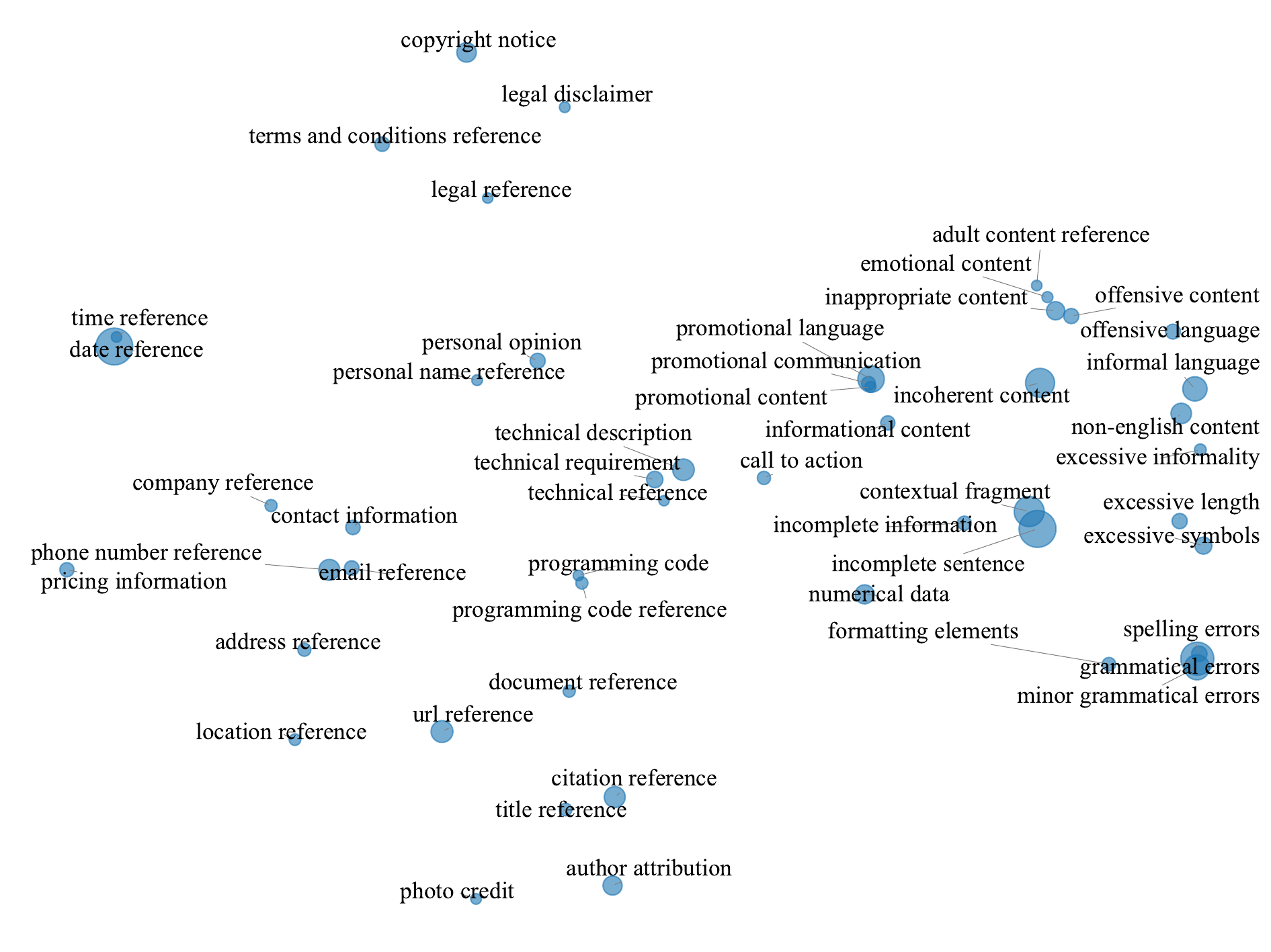}
    \caption{UMAP plot of embeddings of the 50 most frequent LLM-generated label names, created using the Stella-en-400M-v5 model.}
    \label{fig:umap}
\end{figure}

To visualize the distribution of these classes, we generate a 2D UMAP projection \cite{McInnes2018} of the 50 most frequent label embeddings, created using the Stella-en-400M-v5 model \cite{stella} (see also Section \ref{subsec:classifier} below). The UMAP projection reduces the original 1024-dimensional embeddings to 2D, as shown in Figure \ref{fig:umap}, with each dot scaled to represent the relative frequency of each class.

Inspecting the plot, we observe that certain types of low-quality content tend to occupy distinct regions in the space.  For instance, legal texts appear in the top-left, adult and toxic content in the top center-right, and bibliographic references near the bottom. Contact information, such as times, dates, and phone numbers, is loosely grouped on the left, while technical content, like programming code, appears in the center. These patterns suggest that the LLM-generated labels capture meaningful line quality distinctions and form a useful basis for our final class set.

\subsection{Grouping the labels}
\label{subsec:label_grouping}

The next step in our pipeline is to group the 382 detailed labels into a more concise set of broader, more manageable categories, which simplifies training the encoder classifier. We use OpenAI’s o1-preview, a newly released "reasoning" model \cite{openai2024openaio1previewlearningtoreason}, to organize the labels. We instruct the model to create clear, distinct categories that assign each label to only one group. The goal is to produce a set of classes that the classifier can learn and differentiate easily.

\begin{table}[!ht]
    \centering
    \small
    \begin{tabularx}{\linewidth}{lrr}
        \toprule
        \textbf{Category} & \textbf{Lines} & \textbf{\%} \\
        \midrule
        \textit{Clean} & 283,267 & 86.24 \\
        \midrule
        \textit{Formatting, Style \& Errors} & 13,150 & 4.00 \\
        \midrule
        \textit{Bibliographical \& Citation References} & 8,768 & 2.67 \\
        \midrule
        \textit{Promotional \& Spam Content} & 7,339 & 2.23  \\
        \midrule
        \textit{Contact \& Identification Information} & 3,898 & 1.19 \\
        \midrule
        \textit{Navigation \& Interface Elements} & 3,327 & 1.01 \\
        \midrule
        \textit{Technical Specifications \& Metadata} & 3,298 & 1.00 \\
        \midrule
        \textit{Legal \& Administrative Content} & 2,992 & 0.91 \\
        \midrule
        \textit{Offensive or Inappropriate Content} & 2,433 & 0.74 \\
        \midrule
        \textbf{Total} & \textbf{328,472} & \textbf{100} \\
        \bottomrule
    \end{tabularx}
    \caption{Label categories and the number of lines in each category.}
    \label{tab:label_categories}
\end{table}

After manually inspecting the output, we find that the groupings are mostly accurate, though some manual corrections are necessary. For example, the model occasionally fails to assign all labels or places some labels into multiple categories. After fixing these issues, we finalize a classification scheme with 9 broader categories, as shown in Table \ref{tab:label_categories}.

To verify that the labels match human intuition, we conduct a manual inter-annotator agreement (IAA) evaluation on a random sample of 50 documents (726 lines). Two human annotators, familiar with the 9-label class set, assess whether they agree or disagree with the LLM-generated labels. In cases of disagreement, they provide corrected labels. We compute Cohen's Kappa scores comparing human ratings with the LLM's for both the full label set and a simplified binary classification (\textit{Clean} vs. \textit{Non-clean}).

\begin{table}[!ht]
\centering
\small
\begin{tabularx}{\linewidth}{Xrrr}
\toprule
 & \textbf{A1} & \textbf{A2} & \textbf{Avg.} \\ 
\midrule
All labels                    & 0.79  & 0.60  & 0.70  \\ 
\midrule
Clean vs. Non-clean             & 0.78  & 0.67  & 0.73  \\ 
\bottomrule
\end{tabularx}
\caption{Cohen's Kappa scores for human annotators (A1 and A2) vs. the GPT-4o mini generated labels (LLM).}
\label{tab:iaa}
\end{table}

As shown in Table \ref{tab:iaa}, Cohen's Kappa for the full label set is 0.788 for Annotator 1 (A1) and 0.604 for Annotator 2 (A2), with an average of 0.70, indicating moderate to substantial agreement. For the binary classification, Kappa scores improve slightly, with A1 at 0.78 and A2 at 0.67, averaging 0.73. This suggests that while agreement varies, the LLM-based classification generally produces acceptable labels for the FineWeb texts.

These results address RQ1, which examines how well an LLM can identify low-quality content that heuristic filters miss. The LLM's classifications align well with those of human annotators, showing that it succees to detect low-quality lines overlooked by earlier heuristic methods applied to FineWeb data. While there is some variability in the IAA scores, the overall performance supports our LLM-driven approach.

\subsection{Training a classifier} \label{subsec:classifier}

To scale our labeling process for the FineWeb-10BT dataset, we use encoder-based models, which are faster, more cost-effective, and often better suited to classification than large generative LLMs. We experiment with four models: DeBERTa-v3 (base and large variants) \cite{he2021debertav3}, Stella-en-400M-v5 (currently the top model of its size for English text clustering on the MTEB leaderboard \cite{muennighoff2023mtebmassivetextembedding}\footnote{https://huggingface.co/spaces/mteb/leaderboard}), and XLM-RoBERTa-base \cite{DBLP:journals/corr/abs-1911-02116}. The first three models are English-only, while XLM-RoBERTa is multilingual.

For line-by-line classification, we first extract individual lines from the documents, treating each as a separate example. The data is then shuffled and split into training (70\%), development (10\%), and test (20\%) sets using stratification. We add a classification head to each model to generate probabilities across the 9 classes for each line and fine-tune both the classification head and base model. Preliminary tests showed that this approach yielded better results than training only the classification head with a frozen base model.

For training, we use bfloat16 precision, a learning rate of 1e-5, and a batch size of 16. Early stopping is applied with a patience of 5 based on evaluation loss, with a maximum of 5 epochs; however, models typically converge after the first epoch. We also apply label smoothing (0.1) to the cross-entropy loss to improve generalization. Training is done on a single A100 GPU.

\begin{table}[!ht]
    \centering
    \setlength{\tabcolsep}{4pt}
    \small
    \begin{tabularx}{\linewidth}{Xrrrrrr}
        \toprule
        & \textbf{$\mu$ F1} & \textbf{$M$ F1} & \multicolumn{3}{c}{\textbf{Clean}} \\
        \cmidrule(lr){4-6} 
        & & & \textbf{P} & \textbf{R} & \textbf{F1} \\
        \midrule
        DeBERTa-v3-base   & 0.81 & 0.66 & \textbf{0.88} & 0.91 & \textbf{0.90} \\
        DeBERTa-v3-large   & 0.81 & 0.65 & 0.87 & 0.92 & 0.89 \\ 
        Stella-en-400M-v5  & 0.81 & \textbf{0.67} & 0.87 & 0.92 & 0.89 \\
        XLM-RoBERTa-base       & 0.80 & 0.63 & 0.86 & 0.92 & 0.89 \\
        \bottomrule
    \end{tabularx}
    \caption{Comparison of Classifiers on Multiclass Classification using the held-out test set. $\mu$ F1: Micro F1, M F1: Macro F1, P: Precision, R: Recall, F1: F1 score for the \textit{Clean} class.}
    \label{tab:classifier_comparison}
\end{table}

Table \ref{tab:classifier_comparison} presents the evaluation results of the models on the test set. We report micro and macro F1 scores for all classes, along with precision, recall, and F1 for the \textit{Clean} class. The results show that the models perform similarly, with micro F1 scores ranging between 0.80 and 0.81, and macro F1 scores between 0.63 and 0.67. For the \textit{Clean} class, precision ranges from 0.86 to 0.88, recall from 0.91 to 0.92, and F1 between 0.89 and 0.90. These metrics indicate strong performance in distinguishing between high- and low-quality content, though the lower macro F1 score suggests some classes are less easily distinguishable. Additionally, newer or larger models do not significantly improve performance. Thus, for subsequent analyses, we select the DeBERTa-v3-base model.

\begin{figure}[!ht]
    \centering
        \includegraphics[width=\columnwidth]{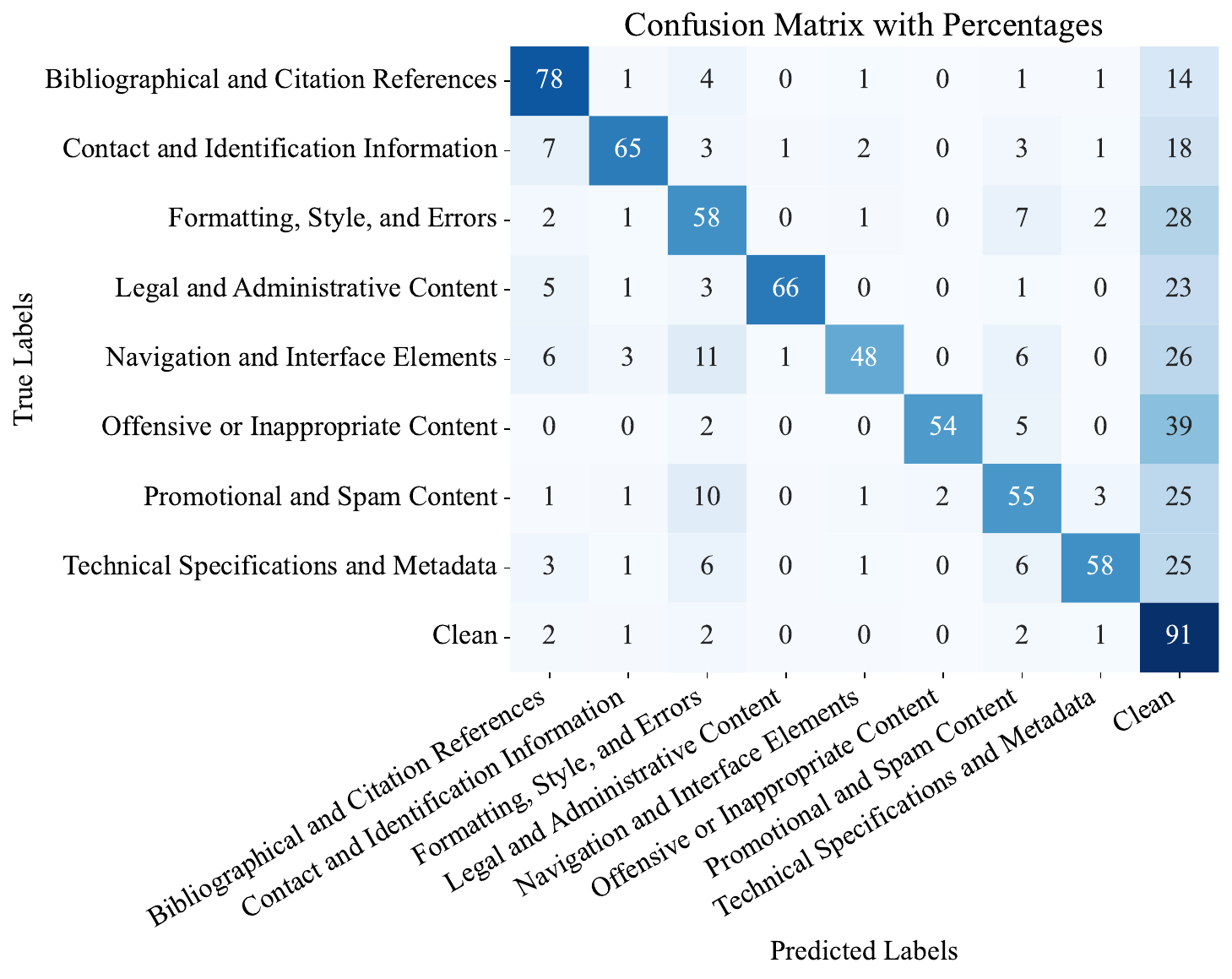}
    \caption{Confusion matrix of predictions from our line quality classifier on the test set.}
    \label{fig:confusion_matrix}
\end{figure}

To further examine the performance of the classifier and spot common misclassifications, we evaluate its predictions on the held-out test set using DeBERTa-v3-base and display the results in a confusion matrix (Figure \ref{fig:confusion_matrix}). Most misclassifications fall into the \textit{Clean} class, indicating strong separation between the other classes. The least distinct class is \textit{Offensive or Inappropriate Content}, likely due to the inherent difficulty in defining clear boundaries for offensive material in LLM training datasets. In contrast, \textit{Bibliographical and Citation References} stands out as the most distinct class, likely due to its easily recognizable formatting and content.

We note that it is preferable for the classifier to err on the side of labeling low-quality lines \textit{Clean} (as shown in the confusion matrix and evaluation scores) rather than mistakenly tagging high-quality lines as low-quality. This bias helps reduce the risk of discarding valuable data from the dataset.

\subsection{Cleaning FineWeb}
\label{subsec:cleaning_fineweb}

Given our classifier's promising evaluation results, we now label the 10B-token subset of FineWeb using our DeBERTa-v3-base classifier. For this task, we simplify to binary classification by focusing only on the probability of the \textit{Clean} class versus all other classes combined, where probabilities closer to 1 indicate high-quality content.

Although the classifier performs well, the \textit{Clean} class makes up 86\% of the data, which may cause the model to produce overconfident predictions for this class. To correct for this imbalance, we apply Platt scaling \cite{platt1999probabilistic} to adjust the predicted probabilities, aiming for a more accurate reflection of the true probability distribution and more reliable thresholding. Specifically, we train a Platt logistic regression model on the held-out test set and apply it on top of the classifier when predicting quality scores for the FineWeb-10BT dataset.

We predict the quality labels for the FineWeb-10BT dataset in shards of 100,000 documents. Within each shard, we process batches of 128 lines, grouping lines by length to speed up processing. We then add a "quality\_score" key to each document, with each item scored from 0 to 1 to four decimal places.

\begin{figure}[!ht]
    \centering
        \includegraphics[width=\columnwidth]{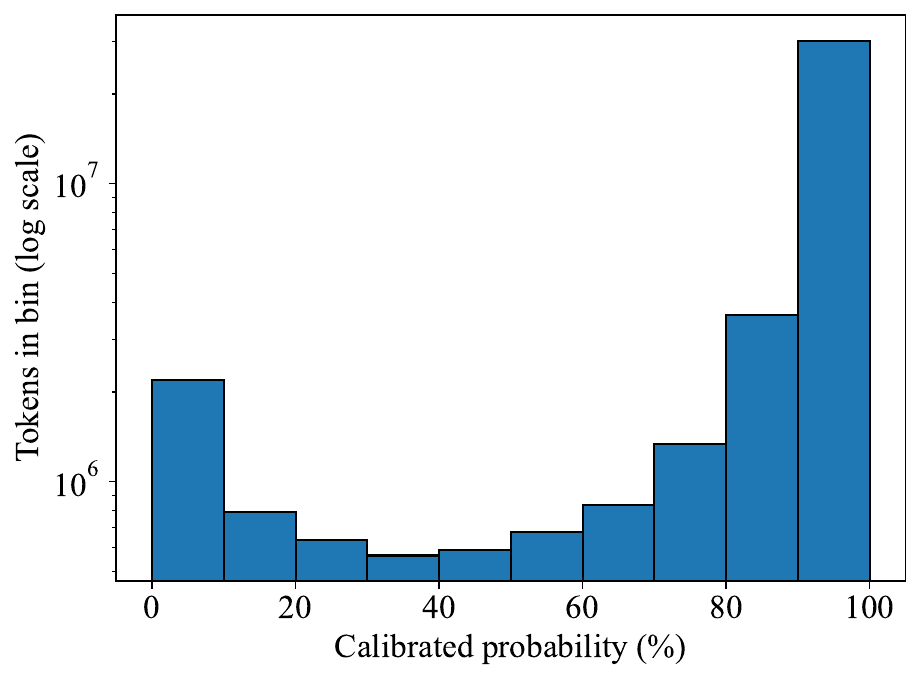}
    \caption{Quality probabilities for a 1M-line sample from FineWeb-10BT, binned in 10\% intervals (log scale). A total of 8\% of lines fall below the 0.50 quality threshold, and 25\% fall below the 0.90 threshold.}
    \label{fig:histogram}
\end{figure}

Figure \ref{fig:histogram} shows a histogram of the quality scores for a 1-million-line sample from FineWeb-10BT, with calibrated probabilities binned in 10\% intervals on a logarithmic scale. The distribution is bimodal, with most lines receiving high-quality scores. About 75\% of lines score above 0.90, while 8\% score below 0.50. Most of the data is concentrated in the highest quality bin (90–100\%), with a smaller cluster confidently assigned very low scores, indicating that the classifier effectively separates high-quality from low-quality lines.

Table \ref{tab:top_and_bottom_clean_lines} shows examples of lines with the highest and lowest quality scores according to our classifier. The highest-scoring lines are coherent, context-rich sentences, while the lowest-scoring lines contain metadata, copyright symbols, tags, and formatting artifacts, demonstrating that the method performs as intended.

\begin{table}[!t]
\centering
\small
\begin{tabularx}{\columnwidth}{>{\raggedright\arraybackslash}Xr}
\toprule
\textbf{Line} & \textbf{Score} \\
\midrule
\multicolumn{2}{l}{\textbf{Lines with highest quality scores}} \\
\midrule
She hopes taking part in the 5K will encourage others to become or stay active. & 0.9674 \\
\midrule
I'd love it if you'd visit and give me your impressions and/or suggestions. & 0.9659 \\
\midrule
We aim to make the ceremony an enjoyable celebration. & 0.9657 \\
\midrule
prayerfully seek peace for our partners in Nigeria. & 0.9655 \\
\midrule
I loved the way this shirt looked and thought it would be cool to wear it. & 0.9655 \\
\midrule
\multicolumn{2}{l}{\textbf{Lines with lowest quality scores}} \\
\midrule
|Also published as||US20040168193| &  0.0057 \\
\midrule
|Tags:||Anglesey, Beach, General, Landscape, Landscape / travel, Lighthouse, Llanddwyn, Sea, Sunrise, Wales, Water| & 0.0056 \\
\midrule
|FOR IMMEDIATE RELEASE||PRESS RELEASE \#MR12-003881| & 0.0055 \\
\midrule
|\textcopyright Sunwest Bank|||||Equal Housing Lender|||||Member FDIC| & 0.0051 \\
\midrule
- \textcopyright - copyright \& copy; or \& \#169; or \& \#xA9; & 0.0050 \\
\midrule
\bottomrule
\end{tabularx}
\caption{Examples of highest and lowest quality lines from a 1M-line FineWeb-10BT sample, with their probabilities of being \textit{Clean}.}
\label{tab:top_and_bottom_clean_lines}
\end{table}

\subsection{Evaluation with GPT-2 and HellaSwag}
\label{subsec:gpt2_eval}

Finally, we evaluate our data cleaning process by pre-training small GPT-2 models (124M parameters) on three versions of the dataset: (1) the original 10B-token sample from FineWeb, (2) a filtered version with a 0.50 quality score threshold, reducing the dataset by 8\%, and (3) a version with a 0.90 quality score threshold, reducing data by 25\%. The training code is adapted from \citet{Khajavi2024}, with modifications specific to our experimental setup.

The models are trained for 18,994 steps (a single epoch on the full FineWeb-10BT dataset) using four A100 GPUs. Every 200 steps, we evaluate model performance on the HellaSwag benchmark \cite{zellers2019hellaswagmachinereallyfinish}, which is widely used to assess the ability of language models to complete sentences in commonsense reasoning contexts. To account for inherent randomness, we repeat the training on all datasets five times each, with each run lasting approximately 5 hours and 30 minutes.

\begin{figure}[!t]
    \centering\includegraphics[width=\columnwidth]{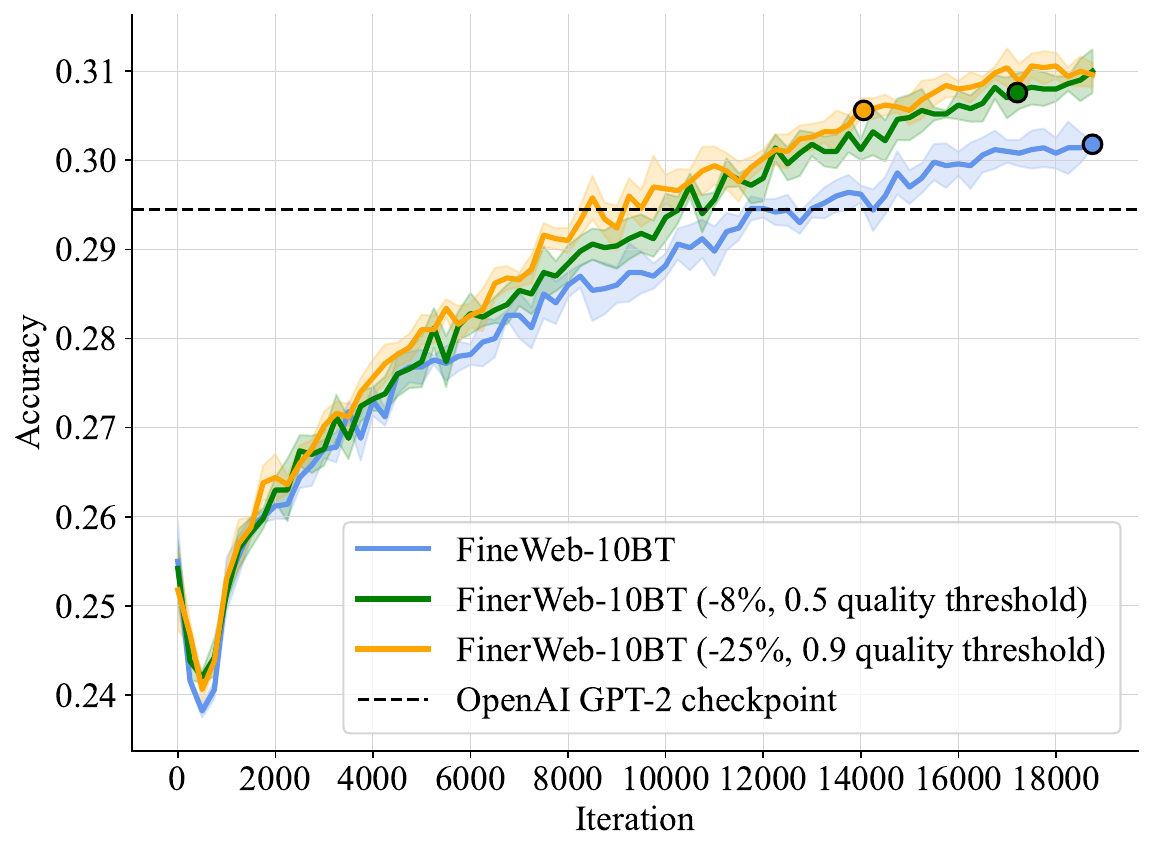}
    \caption{Average HellaSwag accuracy over 5 runs for three models: the original FineWeb-10BT and two cleaned versions with quality thresholds of 0.50 (8\% data reduction) and 0.90 (25\% data reduction). Dot markers indicate epoch ends for each dataset run. GPT-2 (124M) checkpoint accuracy is shown for reference.}
    \label{fig:hellaswag}
\end{figure}

Figure \ref{fig:hellaswag} shows the evaluation results, which indicate a clear positive impact from our data cleaning process. Models trained on the cleaner FinerWeb-10BT datasets---both the 8\% and 25\% reduced versions---consistently outperform those trained on the original FineWeb-10BT data. By the end of 18,994 training steps, both cleaned versions show an average HellaSwag evaluation score that is 0.1 points higher than that of the original dataset. This improvement is robust, as shown by the shaded areas around the lines, representing standard deviations that suggest the effect is unlikely due to random variation across runs.

Additionally, both cleaned models achieve slightly higher HellaSwag accuracy than the original FineWeb-10BT model at their respective epoch ends, as indicated by the colored dots in the plot. Remarkably, both models reach the original dataset’s highest score approximately 6k steps earlier, a 32\% reduction in training time. This means a reduction of roughly 1 hour and 45 minutes, based on our 5 hour 30 minute run time per training round. Interestingly, the 25\% reduced dataset shows a slight edge over the 8\% cleaned data, although the difference is minimal; both clean models ultimately reach an average HellaSwag score of 0.31 within the same number of steps. This suggests that a more aggressive data cleaning strategy could be worth exploring in future work. In summary, our data cleaning process produces models that (1) reach target accuracy faster and (2) achieve higher accuracy within the same training time, addressing our RQ2. 

\section{Discussion} \label{sec:results}

The labels generated by GPT-4o mini reveal both the quantity and types of low-quality lines that remain in FineWeb. The largest categories include lines with grammatical errors, poor formatting, and incomplete sentences, along with recurring items like time stamps, legal jargon, and promotional content. While these elements do not necessarily reduce dataset quality (a good language model should recognize items like copyright notices or phone numbers), our evaluation shows that reducing their prevalence improves both accuracy and training efficiency. These findings suggest that more precise control over the types and proportions of low-quality data included could further benefit model performance. Even when simplified to binary classification, our LLM-driven approach clearly outperforms heuristic methods in enhancing dataset quality.

Specifically, our evaluation on GPT-2 using HellaSwag shows that with less but cleaner data, the model achieves comparable or even slightly better accuracy. While GPT-2 is small relative to SOTA models, our results provide strong evidence that LLM-based data filtering can reduce training time and save energy. Although we tested our method on a small, English-only dataset, this data-driven approach to quality filtering is easily adaptable to other datasets and languages, although low-resource language may suffer from worse LLM performance.

Using an LLM as a judge of text quality introduces some bias, as the model’s training data and design choices influence the resulting labels. For example, mature SOTA LLMs have strong in-built safety features that prevent them from generating harmful or offensive content. In our case, we observe that GPT-4o mini sometimes labels mild expletives, such as "shut up", as toxic, reflecting an overly sensitive filter for offensive language. As described in Sections \ref{subsec:gpt_labeling} and \ref{subsec:label_grouping} we made some manual adjustments to the LLM labeling to account for such biases. Also, the line between low-quality and high-quality is naturally vague, which introduces noise into the data. In future work, we plan to experiment with different models and adjust our prompts to further improve this filtering approach.

\section{Conclusion} \label{sec:conclusion}

In this paper, we propose a novel approach to improving the quality of large-scale language model training datasets through fine-grained, line-level filtering with large language models (LLMs). We first used GPT-4o mini to label a sample from the FineWeb dataset, generating detailed labels that captured low-quality content often overlooked by heuristic filters, addressing our first research question (RQ1). These labels were grouped into broader categories using OpenAI's o1-preview model, followed by training a DeBERTa-v3 classifier to scale the filtering across FineWeb-10BT. Our experiments demonstrate that this LLM-driven filtering pipeline improves model performance (addressing RQ2), as GPT-2 models trained on the filtered dataset achieved higher HellaSwag accuracy with up to 25\% less data than those trained on the original FineWeb-10BT dataset.

These findings suggest that traditional heuristic filters may not be sufficient and that more sophisticated data preprocessing methods are necessary, especially as we face challenges like data scarcity and environmental concerns. Our approach contributes to the emerging field of LLM-based data preprocessing, offering a promising avenue for improving training efficiency and model performance.

In future work, we plan to refine our pipeline by broadening the labeling scheme to provide a more comprehensive description of document contents. We will also experiment with more nuanced filtering approaches, moving beyond simple score-based thresholds, and compare against baselines such as random data reduction to further validate our filtering method. We also plan to test Llama-style models and other architectures to see how our findings scale to newer LLMs. Further evaluations and statistical testing will help strengthen the reliability of our results. Finally, we plan to extend our method to other datasets and languages.

\section*{Acknowledgments}

This project has received funding from the European Union’s Horizon Europe research and innovation programme under Grant agreement No
101070350 and from UK Research and Innovation (UKRI) under the UK government’s Horizon Europe funding guarantee [grant number 10052546]. The contents of this publication are the sole responsibility of its authors and do not necessarily reflect the opinion of the European Union.

This work was supported by the Research Council of Finland. 

Computational resources for this study were provided by CSC --- IT Center for Science.

\bibliographystyle{acl_natbib}
\bibliography{nodalida2025}

\end{document}